# Quantitative Analyses of Chinese Poetry of Tang and Song Dynasties: Using Changing Colors and Innovative Terms as Examples

Chao-Lin Liu
National Chengchi University, Taiwan
chaolin@nccu.edu.tw

**Introduction**

Tang (618-907 AD) and Song (960-1279) dynasties are two very important periods in the development of Chinese literary. The majority forms of the poetry in Tang and Song were Shi (詩) and Ci (詞), respectively. Tang Shi and Song Ci established crucial foundations of the Chinese literature, and their influences in both literary works and daily lives of the Chinese communities last until today.

Recognizing the importance of Tang Shi, a Chinese emperor of the Qing dynasty (1644-1912), Kangxi, ordered to compile a collection of Tang poems, *Quan-Tang-Shi* (*QTS*, 全唐詩). *QTS* contains nearly 50 thousand works of about 2200 poets. A similar effort for compiling a collection of Song Ci from the private sector began in the Ming dynasty (1368-1644), and achieved in a collection called *Quan-Song-Ci* (*QSC*, 全宋詞) in the early Republican period of China (ca. 1937). *QSC* contains around 20 thousand works of about 1330 poets. The exact statistics about *QTS* and *QSC* may vary slightly depending on the sources.

In the past more than a thousand years, literary and linguistic researchers have had done a myriad of research about the poetry of the Tang and Song dynasties. Hence, it is beyond our capacity and not our objective to review the literature in this abstract. Traditional researchers studied and compared poetic works that were produced by different authors and in different time periods to produce insightful and invaluable analyses and commentaries[1]. Most of the time, the researchers focused on the poems of selected poets. Even when computing supports become available, studying poems of specific poets[2] is still an important and popular type of research in poetry.

Software tools facilitate the analysis of poetry from a panoramic perspective, and may lead to applications that would be very challenging in the past. For instance, Zhou[3] and his colleagues analyze the contents of collected couplets and Tang poems for creating couplets. Yan[4] and his colleagues considered topic modeling in automatic composition of Chinese poetry. Lee[5] concentrates on the linguistic analysis and teaching of *QTS*.

---

[1] Tao, Wen-Peng (陶文鵬) (1999) "Research on Tang poems in the first half of the twentieth century," *Journal of Hubei University* (Philosophy and Social Science), v. 5. (in Chinese, http://www.guoxue.com/master/wangpijiang/wpj02.htm)

[2] Jiang, Shao-Yu (蔣紹愚) (2003) "'Moon' and 'Wind' in Li Bai's and Du Fu's poems – Using computers for studying classical poems," *Proc. of the 1st Int'l Conf. on Literature and Information Technologies*. (in Chinese)

[3] Zhou, Ming (周明), Long Jiang, and Jing He (2009) "Generating Chinese couplets and quatrain using a statistical approach," *Proc. of the 23rd Pacific Asia Conf. on Language, Information and Computation*, 43–52.

[4] Yan, Rui (嚴睿), Han Jiang, Mirella Lapata, Shou-De Lin, Xueqiang Lv, and Xiaoming Li (2013) "i, poet: Automatic Chinese poetry composition through a generative summarization framework under constrained optimization," *Proc. of the 23rd Int'l Joint Conf. on Artificial Intelligence*, 2197–2203.

In this presentation, we will discuss some interesting findings in some quantitative analyses of *QTS* and *QSC*.

## Colors and Imageries

Colors are an important ingredient in everyday lives, and actually carried important meanings in religion and social statuses in pre-modern Chinese societies. Wong, a specialist in colors, discusses hidden meanings of colors in China in his book on "The colors of China"[6].

The beauty and imageries conveyed in the poetry originate from the collocations of the written words in the poems. Lo[7] and Huang[8] attempt to classify terms in the poetry by their semantic categories. The results can then serve as a foundation for analyzing the imageries hidden in the poems. Liu[9] and his colleagues emphasize that colors play a crucial role in painting the imaginaries of poems: "colors in poems are like audios in movies", and they analyze the words related to colors in *QTS*.

Using methods for text analysis, one may analyze occurrences and collocations of colors in *QTS* and *QCS*. The main contribution of our work will be illustrating meaningful applications of text analysis for linguistics and literary. The collocations of color words that appeared frequently in *QTS* are certainly interesting[9]. Yet, the analysis can be extended in at least two directions. First, did a poet have specific preferences on some collocations? Second, how were the collocations used by different authors?

Bai Juyi[10] has the largest number of works in *QTS*. He used "白髮"[11] with "青衫"[11], "青雲"[11], "丹砂"[11], and "青山"[11], and "白首"[11] with "青山" and "紅塵"[11] relatively often. Liu Changqin[10], another important poet in the mid Tang period, used "白髮" with "滄洲"[11], and "白首" with "青山", "滄洲", and "青春"[11] relatively often. These different words and collocations convey the imagery of "aging", and the variations in the word choices shed light on the subtle differences between the poets about how they expressed emotions about aging.

---

[5] Lee, John (李思源) (2012) "A classical Chinese corpus with nested part-of-speech tags," *Proc. of the 6th EACL Workshop on Language Technology for Cultural Heritage, Social Sciences, and Humanities*, 75–84.

[6] Wong, Yan T. (黃仁達) (2011) *The Colors of China*, Taipei: Linking Publishing.

[7] Lo, Fengju (羅鳳珠) (2008) "The research of building a semantic category system based on the language characteristic of Chinese poetry," *Proc. of the 9th Cross-Strait Symposium on Library Information Science*. (in Chinese)

[8] Huang, Chu-Ren (黃居仁) (2004) "Text-based construction and comparison of domain ontology: A study based on classical poetry," *Proc. of the 18th Pacific Asia Conf. on Language, Information and Computation*, 17–20.

[9] Liu, Chao-Lin (劉昭麟), Hongsu Wang, Chu-Ting Hsu, Wen-Huei Cheng, and Wei-Yun Chiu (2015) "Color aesthetics and social networks in complete Tang poems: Explorations and discoveries," *Proc. of the 29th Pacific Asia Conference on Language, Information and Computation*, 132–141.

[10] Bai Juyi: 白居易; Liu Changqin: 劉長卿

[11] 白髮: bai2 fa3; 青衫: qing1 shan1; 青雲: qing1 yun1; 丹砂: dan1 sha1; 青山: qing1 shan1; 白首: bai2 shou3; 紅塵: hong2 chen2; 滄洲: cang1 zhou1 ; 青春:qing1 chun1

"白雲"[12] is a very frequent word in *QTS*. Collocating with different words would create different imageries in the poems, e.g., "黃葉"[13], "滄海"[14], "清露"[15], and "流水"[16]. Each of these collocations may brew a different scene in readers' minds, and some of these collocations are more popular than other. It should be interesting for researchers to extract the source poems[17] from the *QTS* to thoroughly study them.

Liu et al. reported that white ("白") is the most frequent color in *QTS*[9]. We may check and find that red ("紅") is the most frequent color in *QCS*. It is interesting to investigate the changes (and their causes) of the popular colors from *QTS* to *QCS*. Again, using text analysis methods, we can find a good approximation of the trend, though obtaining the precise frequencies of the colors requires the techniques of word sense disambiguation (WSD). For instance, "金"[18] could represent a material or a color, so WSD is necessary to achieve precise statistics.

In *QSC*, the most frequent six colors are in the order of "紅", "青", "黃", "綠", "白", and "碧"[19], while, in *QTS*, the most frequent six colors are "白", "青", "紅", "黃", "碧", and "綠". "紅塵", "殘紅", "紅妝", "紅葉", "紅袖", "紅日", and "紅樓"[20] are some of the most frequent red words in *QSC*. The changes in the dominant colors from *QTS* to *QSC* may be a result of the selection process and may be a result of the cultural shift, and is an academically interesting issue to purse further.

**Word Inventions and Influences**

Liu and Wang[21] propose a method to measure and compare the influences of the poems of a poet. Their methods consider whether or not the poems were selected to be included in famous collections.

While our goal is not to challenge Liu and Wang's viewpoint, we would propose to consider also whether poets created new words that were used by later Chinese generations. Isn't it practically meaningful and academically significant to create new words that future generations continue to use? At the time of Tang and Song, poets were at an excellent stage of the Chinese history to achieve such a cultural impact.

---

[12] 白雲: bai yun, white cloud
[13] Collocations of "白雲" and "黃葉" (huang2 ye4, yellow leaves) appeared in poems of 劉長卿(4 times), 盧綸(1), 常袞(1), and 賈島(1).
[14] Collocations of "白雲" and "滄海" (cang1 hai3, broad ocean) appeared in poems of 劉長卿(4), 姚合(1), 崔峒(1), and 賈島(1).
[15] Collocations of "白雲" and "清露" (qing1 lou4, light dew) appeared in poems of 權德輿(1) and 賈島(1).
[16] Collocations of "白雲" and "流水" (liu2 shui3, running water) appeared in poems of 劉禹錫(1), 姚合(1), 皇甫冉(1), 皇甫曾(1), 賈島(1), and 錢起(1).
[17] Two examples by 劉長卿: "白雲留永日，黃葉減餘年" and "近北始知黃葉落，向南空見白雲多".
[18] 金: jin1, gold
[19] 白: bai1, white; 青: qing1, blue; 紅: hong2, red; 黃: huang1, yellow; 碧: bi4, green; 綠: lu4, green
[20] 紅塵: hong2 chen2; 殘紅: can2 hong2; 紅妝: hong2 zhuang1; 紅葉: hong2 ye4; 紅袖: hong2 xui4; 紅日: hong2 ri4; 紅樓: hong2 lou2
[21] Liu, Zunming (劉尊明) and Zhaopeng Wang (2012) *Quantitative Analysis of Ci in Tang and Song Dynasties*, Beijing: Peking University Press.

We conduct an analysis of frequent bigrams in *QTS* and *QSC*, and compare the differences. Words that appeared only in *QSC* are candidates of new words which were invented in Song dynasty. Words that appeared only in *QTS* are candidates of words that failed to survive in Chinese language. Although this process does not really guarantee a water-proof theoretical foundation for word invention, the findings should still serve as a persuasive factor in linguistic, literary, or historical research.

Here are some of such findings. "紅塵"[22] appeared in both *QTS* and *QSC*. "惺忪"[22] is a word that appeared in *QSC*[23] but not in *QTS*, and is still being used in modern Chinese. "空門"[22] is a word that appeared in *QTS* but not in *QSC*, and is being used in Chinese. "武皇"[22] is a word that appeared in *QTS* but not in *QSC*, and is not normally used in Chinese. "酴醾"[24] represents another type of instance. This word appeared much often in *QSC* than in *QTS*.

**Discussions**

A major challenge in analyzing the words in Chinese poetry is word segmentation. Traditional experience indicates that most words in poetry consist of one or two characters[25]. Relying on this heuristic, we can algorithmically analyze the corpora containing about 4.9 million characters at least approximately. To really understand and appreciate the poetry, one should not read them verbatim. Metaphor recognition can be essential for revealing the real intentions of the poets.

Observations resulting from quantitative analysis of *QTS* and *QSC* open windows to promising research opportunities about *QTS*, *QSC*, and their transition. In addition to colors, terms about astronomical objects, floral entities, meteorological phenomena, and geographical sights, are important participants in poetry. Innovative collocations of them paint impressive imageries in readers' minds. Good computational tools can help researchers explore poets' worlds more efficiently.

---

[22] 紅塵: hong2 chen2; 惺忪: xing1 song1; 空門: kong1 men2; 武皇: wu3 huang2
[23] For instance, in〈浣沙溪〉of 周邦彥 we read "薄薄紗廚望似空。簟紋如水浸芙蓉。起来娇眼未惺忪".
[24] 酴醾: tu2 mi2; also written as "酴醾" in *QTS*
[25] Lo, Fengju (羅鳳珠) (2005) "Design and applications of systems for word segmentation and sense classification for Chinese poems," *Proc. of the 4th Conference on Technologies for Digital Archives*. (in Chinese)